\documentclass[11pt,a4paper]{article}
\usepackage[hyperref]{naaclhlt2018}
\usepackage{standalone}
\usepackage{mystuff-nlp}
\usepackage{times}
\usepackage{tikz}
\usepackage{url}
\usepackage{latexsym}
\usepackage{color}
\usepackage{float}
\usepackage{amsfonts,amssymb,amsmath,bm,bbm}
\usepackage{algpseudocode}
\usepackage{graphicx,subfigure}
\usepackage{booktabs}
\usepackage{multirow}

\usepackage{enumitem,array}
\usepackage[titlenumbered,ruled]{algorithm2e}

\usepackage{setspace}

\let\argmax\relax

\DeclareMathOperator*{\argmax}{arg\,max}
\newcommand{\example}[1]{`#1'}
\aclfinalcopy % Uncomment this line for the final submission
\def\aclpaperid{181} %  Enter the acl Paper ID here

\title{Collective Entity Disambiguation with Structured Gradient Tree Boosting}

\author{Yi Yang \quad Ozan Irsoy \quad Kazi Shefaet Rahman\\
	Bloomberg LP\\
	New York, NY 10022\\
	{\tt \{yyang464+oirsoy+krahman7\}@bloomberg.net}
        }

\date{}

\begin{document}
\maketitle
\begin{abstract}
We present a gradient-tree-boosting-based structured learning model for jointly disambiguating named entities in a document. 
Gradient tree boosting is a widely used machine learning algorithm that underlies many top-performing natural language processing systems. %, including named entity linking and co-reference resolution. 
Surprisingly, most works limit the use of gradient tree boosting as a tool for regular classification or regression problems, despite the structured nature of language. 
To the best of our knowledge, our work is the first one that employs the structured gradient tree boosting (SGTB) algorithm for collective entity disambiguation. By defining global features over previous disambiguation decisions and jointly modeling them with local features, our system is able to produce globally optimized entity assignments for mentions in a document.
Exact inference is prohibitively expensive for our globally normalized model. To solve this problem, we propose Bidirectional Beam Search with Gold path (BiBSG), an approximate inference algorithm that is a variant of the standard beam search algorithm. BiBSG makes use of global information from both past and future to perform better local search. 
Experiments on standard benchmark datasets show that SGTB significantly improves upon published results. Specifically, SGTB outperforms the previous state-of-the-art  neural system by near 1\% absolute accuracy on the popular AIDA-CoNLL dataset.\footnote{When ready, the code will be published at \url{https://github.com/bloomberg/sgtb}.}

\end{abstract}

% ***********************************************************
% Introduction
\section{Introduction}
\label{sec:intro}

\begin{figure*}[t!]
\centering
\includegraphics[scale=.28]{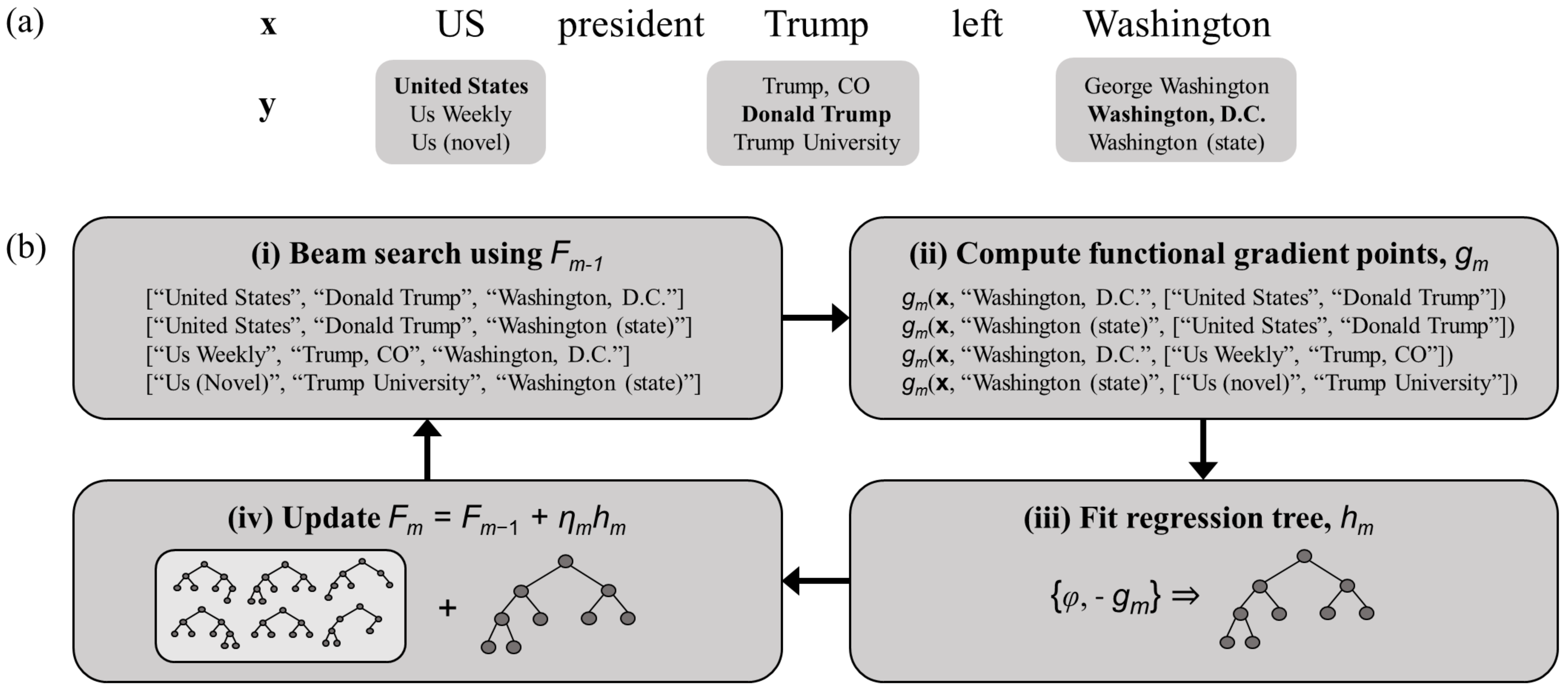}
\caption{(a) Example document $\mathbf{x}$ with entity candidates for each mention (gold entities are in \textbf{bold}); (b) the $m$-th SGTB update iteration: (i) conduct beam search to sample candidate entity sequences (\autoref{sec:train}), (ii) compute point-wise functional gradients for each candidate sequence, (iii) fit a regression tree to the negative functional gradient points with input features, $\phi$, (iv) update the factor scoring function, $F$, by adding the trained regression tree.}
\label{fig:sgb}
\end{figure*}

% Background on NED (task def, importance)
Entity disambiguation (ED) refers to the process of linking an entity mention in a document to its corresponding entity record in a reference knowledge base (e.g., Wikipedia or Freebase). As a core information extraction task, ED plays an important role in the language understanding pipeline, underlying a variety of downstream applications such as relation extraction~\cite{mintz2009distant,riedel2010modeling}, knowledge base population~\cite{ji2011knowledge,dredze2010entity}, and question answering~\cite{berant2013semantic,yih2015semantic}.
This task is challenging because of the inherent ambiguity between mentions and the referred entities. Consider, for example, the mention \example{Washington}, which can be linked to a city, a state, a person, an university, or a lake (\autoref{fig:sgb}).

% Most successful systems are feature based systems
% - characteristics of NED, why statistical features are important (e.g., prior -> 80+ acc)
% - one exception -- EMNLP NN paper, need to pretrain entity embeddings, to implicitly leverage related info, slow
Fortunately, simple and effective features have been proposed to capture the ambiguity that are designed to model the similarity between a mention (and its local context) and a candidate entity, as well as the relatedness between entities that co-occur in a single document. These are typically statistical features estimated from entity-linked corpora, and similarity features that are pre-computed using distance metrics such as cosine. For example, a key feature for ED is the \emph{prior probability} of an entity given a specific mention, which is estimated from mention-entity co-occurrence statistics. This simple feature alone can yield 70\% to 80\% accuracy on both news and Twitter texts~\cite{lazic2015plato,guo2013link}.

% Gradient tree boosting has been shown to perform well on the task (cite CoNLL paper, my ACL paper), due to non-linear model etc.
% However, no structured gradient tree boosting that captures entity-entity information
To capture the non-linear relationships between the low-dimensional dense features like statistical features, sophisticated machine learning models such as neural networks and gradient tree boosting are preferred over linear models. In particular, gradient tree boosting has been shown to be highly competitive for ED in recent work~\cite{yang2015smart,yamada2016joint}.
%Compared to other popular non-linear models, tree-based models can handle categorical features and count data better. 
However, although achieving appealing results, existing gradient-tree-boosting-based ED systems typically operate on each individual mention, without attempting to jointly resolve entity mentions in a document together. Joint entity disambiguation has been shown to significantly boost performance when used in conjunction with other machine learning techniques~\cite{ratinov2011local,hoffart2011robust}. However, how to train a global gradient tree boosting model that produces coherent entity assignments for all the mentions in a document is still an open question.
%Collective disambiguation benefits ED clues that; or employ heuristics to leverage . 

% This work: structured gradient tree boosting to model entity-entity coherence, inspired by globally normalized training of NN, we proposed a globally normalized gradient tree boosting model
In this work, we present, to the best of our knowledge, the first structured gradient tree boosting (SGTB) model for collective entity disambiguation. Building on the general SGTB framework introduced by~\newcite{yang2015smart}, we develop a globally normalized model for ED that employs a conditional random field (CRF) objective~\cite{lafferty2001conditional}. The model permits the utilization of global features defined between the current entity candidate and the entire decision history for previous entity assignments, which enables the global optimization for all the entity mentions in a document. As discussed in prior work~\cite{smith2007weighted,andor2016globally}, globally normalized models are more expressive than locally normalized models.
%As shown in the work by~\newcite{andor2016globally}, globally normalized models are strictly more expressive than locally normalized models.

% Training: bidirectional beam search + distributed training
%Global models often suffer from the difficulty of computing the partition function (normalization term), which can be addressed by the beam search approximation, in which we keep track of multiple hypotheses and sum over the paths in the beam. 
As in many other global models, our SGTB model suffers from the difficulty of computing the partition function (normalization term) for training and inference.  We adopt beam search to address this problem, in which we keep track of multiple hypotheses and sum over the paths in the beam. In particular, we propose Bidirectional Beam Search with Gold path (BiBSG) technique that is specifically designed for SGTB model training. Compared to standard beam search strategies, BiBSG reduces model variance and also enjoys the advantage in its ability to consider both past and future information when predicting an output.
%we need to compute point-wise functional gradients for estimating the auxiliary regression models. We propose Beam Search with Gold path (BSG) training technique to calculate the gradients at each time step of a training sequence.
%We can adopt standard beam search training techniques such as early update~\cite{collins2004incremental} and LaSO~\cite{daume2005learning} to calculate the gradients at the last time step of a training sequence or when the gold path falls out the beam. However, our preliminary studies suggest that these methods often yield unsatisfactory performance. There is typically a small amount of training data available for ED, which limits the number of functional gradient points generated by these training strategies.  To tackle this problem, we propose Beam Search with Gold path (BSG), which keeps the gold path always in the beam, and collect functional gradient points at each time step of a training sequence. 
%In addition, we adopt the recently proposed Bidirectional Beam Search (BiBS)~\cite{sun2017bidirectional} that enjoys the advantage over standard beam search in its ability to consider both past and future information when predicting an output. Our full model, Structured Gradient Tree Boosting trained with Bidirectional Beam Search with Gold path (SGTB-BiBSG), outperfoms best published results on a variety of standard ED datasets.
%achieves state-of-the-art results on a variety of standard entity disambiguation datasets.

% contributions
% Results: strong
Our contributions are:
\begin{itemize}
    \item We propose a SGTB model for collectively disambiguating entities in a document. By jointly modeling local decisions and global structure, SGTB is able to produce globally optimal entity assignments for all the mentions.
    \item We present BiBSG, an efficient algorithm for approximate bidirectional inference. The algorithm is tailored to SGTB models, which can reduce model variance by generating more point-wise functional gradients for estimating the auxiliary regression models.
    \item SGTB achieves state-of-the-art (SOTA) results on various popular ED datasets, and it outperforms the previous SOTA systems by $1$-$2\%$ absolute accuracy on the AIDA-CoNLL~\cite{hoffart2011robust}  dataset.
    % We conduct extensive experiments on popular entity disambiguation datasets in both in-domain and cross-domain settings. 
\end{itemize}

% ***********************************************************
% Model
\section{Model}
\label{sec:model}

In this section, we present a SGTB model for collective entity disambiguation. We first formally define the task of ED, and then describe a structured learning formalization for producing globally coherent entity assignments for mentions in a document. Finally, we show how to optimize the model using functional gradient descent.

% entity disambiguation problem (TODO: find a running example, and draw a figure with it. Then use the example here)
For an input document, assume that we are given all the mentions of named entities within it. Also assume that we are given a lexicon that maps each mention to a set of entity candidates in a given reference entity database (e.g., Wikipedia or Freebase).
%Assume we are given a reference entity database (e.g., Wikipedia or Freebase), and a lexicon that maps a mention into a set of entity candidates. For an input document, we also assume that all the mentions of named entities in the document are given. 
The ED system maps each mention in the document to an entry in the entity database. Since a mention is often ambiguous on its own (i.e., the lexicon maps the mention to multiple entity candidates), the ED system needs to leverage two types of contextual information for disambiguation: local information based on the entity mention and its surrounding words, and global information that exploits the document-level coherence of the predicted entities. Note that modeling entity-entity coherence is very challenging, as the long-range dependencies between entities correspond to exponentially large search space. 

% Model formalization, different from S-MART, need to capture long-range dependency
We formalize this task as a structured learning problem. Let $\mathbf{x}$ be a document with $T$ target mentions, and $\mathbf{y} = \{ y_t \}_{t=1}^T$ be the entity assignments of the mentions in the document. We use $S(\mathbf{x}, \mathbf{y})$ to denote the joint scoring function between the input document and the output structure.  In traditional NLP tasks, such as part-of-speech tagging and named entity recognition, we often rely on low-order Markov assumptions to decompose the global scoring function into a summation of local functions. ED systems, however, are often required to model nonlocal phenomena, as any pair of entities is potentially interdependent. Therefore, we choose the following decomposition:
\begin{equation}
S (\mathbf{x}, \mathbf{y}) = \sum^T_{t=1} F (\mathbf{x}, y_t, \mathbf{y}_{1:t-1}),
\label{eq:joint-score}
\end{equation}
where $F (\mathbf{x}, y_t, \mathbf{y}_{1:t-1})$ is a factor scoring function. Specifically, a local prediction $y_t$ depends on all the \emph{previous decisions}, $\mathbf{y}_{1:t-1}$ in our model, which resembles recurrent neural network (RNN) models~\cite{elman1990finding,hochreiter1997long}.

% CRF loss not MEMM, because globally normalized models much better
We adopt a CRF loss objective, and define a distribution over possible output structures as follows:
\begin{equation}
    p (\mathbf{y} | \mathbf{x}) = \frac{\exp \{ \sum^T_{t=1} F (\mathbf{x}, y_t, \mathbf{y}_{1:t-1}) \}}{Z (\mathbf{x})},
    \label{eq:log-linear-model}
\end{equation}
where 
\begin{equation*}
Z (\mathbf{x}) = \sum_{\mathbf{y}' \in Gen(\mathbf{x})} \exp \{ \sum^T_{t=1} F (\mathbf{x}, y'_t, \mathbf{y}'_{1:t-1}) \}
\end{equation*}
and $Gen(\mathbf{x})$ is the set of all possible sequences of entity assignments depending on the lexicon. $Z (\mathbf{x}) $ is then a global normalization term. As shown in previous work, globally normalized models are very expressive, and also avoid the label bias problem~\cite{lafferty2001conditional,andor2016globally}. The inference problem is to find
\begin{equation}
\argmax_{\mathbf{y} \in Gen(\mathbf{x})} p (\mathbf{y} | \mathbf{x}) = \argmax_{\mathbf{y} \in Gen(\mathbf{x})} \sum^T_{t=1} F (\mathbf{x}, y_t, \mathbf{y}_{1:t-1}).
\end{equation}

\subsection{Structured gradient tree boosting}

An overview of our SGTB model is shown in~\autoref{fig:sgb}. The model minimizes the negative log-likelihood of the data,
\begin{equation}
\begin{aligned}
L (\mathbf{y}^*, S (\mathbf{x}, \mathbf{y}))  &= - \log p (\mathbf{y}^* | \mathbf{x}) \\
                                                  &= \log Z(\mathbf{x}) - S(\mathbf{x}, \mathbf{y}^*),
\end{aligned}
\label{eq:loss}
\end{equation}
where $\mathbf{y}^*$ is the gold output structure.

In a standard CRF, the factor scoring function is typically assumed to have this form: $F (\mathbf{x}, y_t, \mathbf{y}_{1:t-1}) = \mathbf{\theta}^\top \phi (\mathbf{x}, y_t, \mathbf{y}_{1:t-1})$, where $\phi (\mathbf{x}, y_t, \mathbf{y}_{1:t-1})$ is the feature function and $\mathbf{\theta}$ are the model parameters. 
The key idea of SGTB is that, instead of defining a parametric model and optimizing its parameters, we can directly optimize the factor scoring function $ F (\cdot)$ iteratively by performing gradient descent in function space. In particular, suppose $F (\cdot)=F_{m-1} (\cdot)$ in the $m$-th iteration, we will update $F (\cdot)$ as follows:
\begin{equation}
\begin{aligned}
F_m(\mathbf{x}, y_t, \mathbf{y}_{1:t-1}) &= F_{m-1}(\mathbf{x}, y_t, \mathbf{y}_{1:t-1}) \\
&- \eta_m g_m (\mathbf{x}, y_t, \mathbf{y}_{1:t-1}),
\end{aligned}
\end{equation}
where  
\begin{equation}
\begin{aligned}
g_m (\mathbf{x}, y_t, &\mathbf{y}_{1:t-1}) = \frac{\partial L (\mathbf{y}^*, S (\mathbf{x}, \mathbf{y}))}{\partial F(\mathbf{x}, y_t, \mathbf{y}_{1:t-1}))} \\
&= p(\mathbf{y}_{1:t} | \mathbf{x}) - \mathbf{1}[\mathbf{y}_{1:t} = \mathbf{y}^*_{1:t}]
\end{aligned}
\label{eq:func-grad}
\end{equation}
is the functional gradient, $\eta_m$ is the learning rate, and $\mathbf{1} [\cdot]$ represents an indicator function, which returns $1$ if the predicted sequence matches the gold one, and $0$ otherwise. We initialize $F (\cdot)$ to $0$ ($F_0 (\cdot) = 0$).

We can approximate the negative functional gradient $-g_m(\cdot)$ with a regression tree model $h_m(\cdot)$ by fitting the training data $\{ \phi (\mathbf{x}^{(i)}, y_t^{(i)}, \mathbf{y}_{1:t-1}^{(i)}) \}$ to the point-wise negative functional gradients (also known as residuals) $\{ -g_m (\mathbf{x}^{(i)}, y_t^{(i)}, \mathbf{y}_{1:t-1}^{(i)}) \}$. Then the factor scoring function can be obtained by
\begin{equation}
F (\mathbf{x}, y_t, \mathbf{y}_{1:t-1}) = \sum_{m=1}^M \eta_m h_m (\mathbf{x}, y_t, \mathbf{y}_{1:t-1}) ,
\end{equation}
where $h_m (\mathbf{x}, y_t, \mathbf{y}_{1:t-1})$ is called a basis function. We set $\eta_m=1$ in this work.

% ***********************************************************
% Training
% need a diagram to describe the algo

\section{Training}
\label{sec:train}

%As mentioned before, the search space for the output sequences $\{ \mathbf{y}_{1:t}^{(i)} \}$ given a document $\mathbf{x}^{(i)}$ is exponentially large. We will discuss how to sample the output sequences to train SGTB using beam search in~\autoref{sec:train}. 

Training the SGTB model requires computing the point-wise functional gradients with respect to training documents and candidate entity sequences. This is challenging, due to the exponential output structure search space.  First, we are not able to enumerate all possible candidate entity sequences. Second, computing the conditional probabilities shown in~\autoref{eq:func-grad} is intractable, as it is prohibitively expensive to compute the partition function $Z(\mathbf{x})$ in~\autoref{eq:log-linear-model}. Beam search can be used to address these problems. We can compute point-wise functional gradients for candidate entity sequences in the beam, and approximately compute the partition function by summing over the elements in the beam.

%as shown in~\autoref{eq:func-grad}. Exact inference is intractable, due to the exponential output structure search space. Beam search can be used to approximately compute the partition function $Z(\mathbf{x})$ (\autoref{eq:log-linear-model}) by summing over the elements in the beam. 

In this section, we present a bidirectional beam search training algorithm that always keeps the gold sequence in the beam. The algorithm is tailored to SGTB, and improves standard training methods in two aspects: (1) it reduces model variance by collecting more point-wise function gradients to train a regression tree; (2) it leverages information from both past and future to conduct better local search.

% early update or LaSO is the standard
% We turn to gold path
\subsection{Beam search with gold path}
\label{sec:model:bsg}

The early update~\cite{collins2004incremental} and LaSO~\cite{daume2005learning, xu2007learning} strategies are widely adopted with beam search for updating model parameters in previous work. Both methods keep track of the location of the gold path in the beam while decoding a training sequence. A gradient update step will be taken if the gold path falls out of the beam at a specific time step $t$ or after the last step $T$. Adapting the strategies to SGTB training is straightforward. We will compute point-wise functional gradients for all candidate entity sequences after time step $T$ or when the gold sequence falls out the beam.
%Early update and LaSO differ primarily in that the former discards a training example after the first search error, whereas LaSO resumes searching after an error from a state that includes the gold partial structure. 
Both early update and LaSO are typically applied to online learning scenarios, in which model parameters are updated after passing one or a few training sequences.

% fit paradigm belong? need better wording
SGTB training, however, fits the batch learning paradigm. In each training epoch, a SGTB model will be updated only once using the regression tree model fit on the point-wise negative functional gradients. The gradients are calculated with respect to the output sequences obtained from beam search.
%in every training epoch  we calculate functional gradient points for all the training instances; fit a regression tree using the point-wise functional gradients; and update the model by adding the regression tree to the model. 
We propose a simple training strategy that computes and collects point-wise functional gradients at every step of a training sequence. In addition, instead of passively monitoring the gold path, we always keep the gold path in the beam to ensure that we have valid functional gradients at each time step. The new beam search training method, Beam Search with Gold path (BSG), generates much more point-wise functional gradients than early update or LaSO, which can reduce the variance of the auxiliary regression tree model. As a result, SGTB trained with BSG consistently outperforms early update or LaSO in our exploratory experiments, and it also requires fewer training epochs to converge.\footnote{Early update and LaSO perform similarly, thus we only report results for early update in~\autoref{sec:exp}.}

% bidirectional
\subsection{Bidirectional beam search}

During beam search, if we consider a decision made at time step $t$, the joint probability $p (\mathbf{y} | \mathbf{x})$ can be factorized around $t$ as follows:
\begin{equation}
\begin{aligned}
p (\mathbf{y} | \mathbf{x}) = p & (\mathbf{y}_{1:t-1} | \mathbf{x}) \cdot p(y_t | \mathbf{y}_{1:t-1}, \mathbf{x})  \\
                            \cdot & p(\mathbf{y}_{t+1:T} | y_t, \mathbf{y}_{1:t-1}, \mathbf{x}).
\end{aligned}
\label{eq:joint}
\end{equation}

Traditional beam search performs inference in a unidirectional (left-to-right) fashion. Since the beam search at time step $t$ considers only the beam sequences that were committed to so far, $\{\mathbf{y}_{1:t-1}\}$, it effectively approximates the above probability by assuming that all futures are equally likely, i.e. $p(\mathbf{y}_{t+1:T} | y_t, \mathbf{y}_{1:t-1}, \mathbf{x})$ is uniform. Therefore, at any given time, there is no information from the future when incorporating the global structure.

In this work, we adopt a Bidirectional Beam Search (BiBS) methodology that incorporates multiple beams to take future information into account~\cite{sun2017bidirectional}. It makes two simplifying assumptions that better approximate the joint probability above while remaining tractable: (1) future predictions are independent of past predictions given $y_t$; (2) $p(y_t)$ is uniform. These yield the following approximation:
\begin{equation}
\begin{aligned}
  p (\mathbf{y}_{t+1:T} | y_t, \mathbf{y}_{1:t-1}, \mathbf{x})
                    &= p (\mathbf{y}_{t+1:T} | y_t, \mathbf{x}) \\
                    \propto p(y_t  | \mathbf{y}_{t+1:T}, & \mathbf{x}) \cdot
                               p(\mathbf{y}_{t+1:T} |  \mathbf{x}).
\end{aligned}
\end{equation}
Substituting this back into~\autoref{eq:joint} therefore yields:
\begin{equation}
\begin{aligned}
p (\mathbf{y} | \mathbf{x})  \propto p & (\mathbf{y}_{1:t-1} | \mathbf{x}) \cdot p(y_t | \mathbf{y}_{1:t-1}, \mathbf{x}) \\
                             \cdot & p(y_t | \mathbf{y}_{t+1:t}, \mathbf{x}) \cdot p(\mathbf{y}_{t+1:T} | \mathbf{x}),
\end{aligned}
\label{eq:joint-appr}
\end{equation}
which decomposes into multiplication of a forward probability and a backward probability. In~\cite{sun2017bidirectional}, these are retrieved from forward and backward recurrent networks, whereas in our work we use the joint scores (log probabilities shown in~\autoref{eq:joint-score}) computed for partial sequences from forward and backward beams.

\begin{algorithm}[h]
  \small
  \setstretch{1.1}
  \SetKwInOut{Input}{Input}
  \SetKwInOut{Output}{Output}
  \Input{input document $\mathbf{x}$, candidate sequences $\{ \mathbf{y} \}$, \newline joint scoring function $S(\mathbf{x}, \mathbf{y}_{t_1:t_2})$}
  \Output{beam sequence set $C$}
 $C \leftarrow \emptyset$\\
 \While{not converged}{
  // \texttt{forward beam search}\\
  \For{$t = 1, \cdots, T$}{
   $C^{(F)} \leftarrow \text{top-B}_{\mathbf{y}_{1:t}} [S(\mathbf{x}, \mathbf{y}_{1:t}) + S(\mathbf{x}, \mathbf{y}_{T:t})]$\\
   // \texttt{add gold subsequence}\\
   $C^{(F)} \leftarrow C^{(F)} \cup \{ \mathbf{y}_{1:t}^* \}$\\
   $C \leftarrow C \cup C^{(F)}$
   }
   // \texttt{backward beam search}\\
   \For{$t = T, \cdots, 1$}{
   $C^{(B)} \leftarrow \text{top-B}_{\mathbf{y}_{T:t}} [S(\mathbf{x}, \mathbf{y}_{T:t}) + S(\mathbf{x}, \mathbf{y}_{1:t})]$\\
   // \texttt{add gold subsequence}\\
   $C^{(B)} \leftarrow C^{(B)} \cup \{ \mathbf{y}_{T:t}^* \}$\\
   $C \leftarrow C \cup C^{(B)}$
  }
 }
 \caption{Bidirectional Beam Search with Gold path (BiBSG)}
 \label{eq:bibsg}
\end{algorithm}

The full inference algorithm, Bidirectional Beam Search with Gold path (BiBSG), is presented in~\autoref{eq:bibsg}. When performing the forward pass to update the forward beam, forward joint scores, $S(\mathbf{x}, \mathbf{y}_{1:t})$, are computed with respect to current forward beam, and backward joint scores, $S(\mathbf{x}, \mathbf{y}_{T:t})$, are computed with respect to previous backward beam. A similar procedure is used for the backward pass. The search converges very fast, and we use two rounds of bidirectional search as a good approximation.
%BiBS then proceeds by alternating between forward and backward beam searches with respect to the approximate joint probability (\autoref{eq:joint2}). W.L.O.G., when performing the forward pass, forward probabilities,  $p(\mathbf{y}_{1:t-1} | \mathbf{x}) \cdot p(y_t | \mathbf{y}_{1:t-1}, \mathbf{x})$, are computed w.r.t current forward beam and backward probabilities, $p(y_t | \mathbf{y}_{t+1:T}, \mathbf{x}) \cdot p(\mathbf{y}_{t+1:T} | \mathbf{x})$, are computed w.r.t. previous backward beam. Therefore, beam expansion considers $B \times |\mathcal{Y}_t| \times B$ instances and the top $B$ of them is used to update the current forward beam. In this work we use two rounds (pairs of forward and backward passes) of bidirectional search. In order to train a regression tree for SGTB, Bidirectional Beam Search with Gold path (BiBSG) again keeps gold sequences in the beams and computes point-wise functional gradients for all the partial sequences in both forward and backward beams, $\{ \mathbf{y}_{1:t} \}_{t=1}^T$ and $\{ \mathbf{y}_{T:t} \}_{t=T}^1$.
Finally, SGTB-BiBSG compares the conditional probabilities $p(\mathbf{y}^{\text{($\cdot$)}} | \mathbf{x})$ of the best scoring output sequences $\mathbf{y}^{\text{(F)}}$ and $\mathbf{y}^{\text{(B)}}$ obtained from the forward and backward beams. The final prediction is the sequence with the higher conditional probability score.

% ***********************************************************
% Implementation
\section{Implementation}
\label{sec:imple}

We provide implementation details of our SGTB systems, including entity candidate generation, adopted local and global features, and some efforts to make training and inference faster.

\subsection{Candidate selection}
\label{sec:imple:cand}

We use a mention prior $\hat{p}(y | x)$ to select entity candidates for a mention $x$. Following~\newcite{ganea2017deep}, the prior is computed by averaging mention prior probabilities built from mention-entity hyperlink statistics from Wikipedia\footnote{We use a Wikipedia snapshot as of Feb. 2017.} and a large Web corpus~\cite{spitkovsky2012cross}. Given a mention, we select the top $30$ entity candidates according to $\hat{p}(y | x)$.  

%Coreference resolution systems have been exploited to help to improve entity candidate generation in previous work~\cite{pershina2015personalized,lazic2015plato}. For a group of mentions that refer to the same entity, we may select entity candidates based on the one with the least ambiguity. Instead of using complex coreference systems, 
We also use a simple heuristic proposed by~\newcite{ganea2017deep} to improve candidate selection for persons: for a mention $x$, if there are mentions of persons that contain $x$ as a continuous subsequence of words, then we consider the candidate set obtained from the longest mention for the mention $x$.

\subsection{Features}
The feature function $\phi (\mathbf{x}, y_t, \mathbf{y_{1:t-1}})$ can be decomposed into the summation of a local feature function $\phi_L (\mathbf{x}, y_t)$ and a global feature function $\phi_G (y_t, \mathbf{y}_{1:t-1})$.

\paragraph{Local features}
We consider standard local features that have been used in prior work, including mention priors $p(y | x)$ obtained from different resources; entity popularity features based on Wikipedia page view count statistics;\footnote{We obtain the statistics of Feb. 2017 and Dec. 2011 from \url{https://dumps.wikimedia.org/other/pagecounts-ez/merged/} .} named entity recognition (NER) type features given by an in-house NER system trained on the CoNLL 2003 NER data~\cite{tjong2003introduction}; entity type features based on Freebase type information; and three textual similarity features proposed by~\newcite{yamada2016joint}.\footnote{We obtain embeddings jointly trained for words and entities from~\cite{ganea2017deep}.}

\iffalse
\begin{itemize}
    \item \textbf{Mention priors}: we use mention priors $p(e | m)$ obtained from different resources as separated features.
    \item \textbf{Entity popularity}: we adopt two entity popularity features based on Wikipedia page view count statistics.\footnote{We obtain the statistics of Feb. 2017 and Dec. 2011 from \url{https://dumps.wikimedia.org/other/pagecounts-ez/merged/} .}
    \item \textbf{NER types}: we include four named entity recognition (NER) type features given by an in-house NER system trained on the CoNLL 2003 NER data~\cite{tjong2003introduction}.
    \item \textbf{Entity types}: we map each Wikipedia page to one of the four NER types based on Freebase type information and Wikipedia to Freebase mappings.
    \item \textbf{Textual similarity}: we use three textual similarity features based on embeddings jointly trained for words and entities~\cite{ganea2017deep}. Following~\newcite{yamada2016joint}, we derive three textual vectors by averaging the embeddings of words of a mention $m$, of the sentence containing $m$, and of the document respectively. Then we consider three features that are cosine similarities between the textual vectors and the entity vector.
\end{itemize}
 \fi
   
\paragraph{Global features}
Three features are utilized to characterize entity-entity relationships: entity-entity co-occurrence counts obtained from Wikipedia, and two cosine similarity scores between entity vectors based on entity embeddings from~\cite{ganea2017deep} and Freebase entity embeddings released by Google\footnote{\url{https://code.google.com/archive/p/word2vec/}} respectively. We denote the entity-entity features between entities $y_t$ and $y_{t'}$  as $\phi_E (y_t, y_{t'})$.

At step $t$ of a training sequence, we quantify the coherence of $y_t$ with respect to previous decisions $\mathbf{y}_{1:y-1}$ by first extracting entity-entity features between $y_t$ and $y_{t'}$ where $1 \leq t' \leq t-1$, and then aggregating the information to have a global feature vector $\phi_G (y_t, \mathbf{y}_{1:t-1})$ of a fixed length:
\begin{align*}
\phi_G (y_t, \mathbf{y}_{1:t-1}) =& \sum_{t'=1}^{t-1} \frac{\phi_E (y_t, y_{t'})} {t-1} \\
&\oplus \max_{t'=1}^{t-1} \phi_E (y_t, y_{t'}),
\end{align*}
where $\oplus$ denotes concatenation of vectors.

\subsection{Efficiency}

Global models are powerful and effective, but often at a cost of efficiency. We discuss ways to speed up training and inference for SGTB models. 
%primarily from three aspects: fast feature extraction, partial structure score caching, and parallelization. 

Many of the adopted features such as mention priors and entity-entity co-occurrences can be extracted once and retrieved later with just a hash map lookup. The most expensive features are the cosine similarity features based on word and entity embeddings. By normalizing the embeddings to have a unit norm, we can obtain the similarity features using dot products. We find this simple preprocessing makes feature extraction faster by two orders of magnitude.

%partial structure score caching

SGTB training can be easily parallelized, as the computation of functional gradients are independent for different documents. During each training iteration, we randomly split training documents into different partitions, and then calculate the point-wise functional gradients for documents of different partitions in parallel.

% ***********************************************************
% Experiment
\section{Experiments}
\label{sec:exp}

In this section, we evaluate SGTB on some of the most popular datasets for ED. After describing the experimental setup, we compare SGTB with previous state-of-the-art (SOTA) ED systems and present our main findings in~\autoref{sec:exp:results}.

\subsection{Data}

\begin{table} [t!]
\centering
\small
%\addtolength{\tabcolsep}{-2pt}
\begin{tabular}{llll}
    \toprule
    Dataset & \# mention & \# doc & \parbox{1.6cm}{\# mention \\ per doc}  \\ \midrule
    AIDA-train & 18,448 & 946 &19.5  \\
    AIDA-dev & 4,791 & 216 & 22.1 \\
    AIDA-test & 4,485 & 231 & 19.4 \\ \midrule
    AQUAINT & 727 & 50 & 14.5 \\
    MSNBC & 656 & 20 & 32.8 \\
    ACE & 257 & 36 & 7.1 \\ \midrule
    CWEB & 11,154 & 320 & 34.8 \\
    WIKI & 6,821 & 320 & 21.3 \\
    \bottomrule
\end{tabular}
\caption{Statistics of the ED datasets used in this work.}
\label{tab:data}
\end{table}

We use six publicly available datasets to validate the effectiveness of SGTB. AIDA-CoNLL~\cite{hoffart2011robust} is a widely adopted dataset for ED based on the CoNLL 2003 NER dataset~\cite{tjong2003introduction}. It is further split into training (AIDA-train), development (AIDA-dev), and test (AIDA-test) sets.\footnote{AIDA-dev and AIDA-test are also referred as AIDA-a and AIDA-b datasets in previous work.} AQUAINT~\cite{milne2008learning}, MSNBC~\cite{cucerzan2007large}, and ACE~\cite{ratinov2011local} are three datasets for Wikification, which also contain Wikipedia concepts beyond named entities. These datasets were recently cleaned and updated by~\newcite{guo2016robust}. WIKI and CWEB are automatically annotated datasets built from the ClueWeb and Wikipedia corpora by~\newcite{guo2016robust}. The statistics of these datasets are available in~\autoref{tab:data}. 

\subsection{Experimental settings}

Following previous work~\cite{guo2016robust,ganea2017deep}, we evaluate our models on both \emph{in-domain} and \emph{cross-domain} testing settings. In particular, we train our models on AIDA-train set, tune hyperparameters on AIDA-dev set, and test on AIDA-test set (in-domain testing) and all other datasets (cross-domain testing). We follow prior work and report in-KB accuracies for AIDA-test and Bag-of-Title (BoT) F1 scores for the other test sets.

Two AIDA-CoNLL specific resources have been widely used in previous work. In order to have fair comparisons with these works, we also adopt them only for the AIDA datasets. First, we use a mention prior obtained from aliases to candidate entities released by~\newcite{hoffart2011robust} along with the two priors described in~\autoref{sec:imple:cand}. Second, we also experiment with PPRforNED, an entity candidate selection system released by~\newcite{pershina2015personalized}. It is unclear how candidates were pruned, but the entity candidates generated by this system have high recall and low ambiguity, and they contribute to some of the best results reported for AIDA-test~\cite{yamada2016joint,sil2018neural}.

%\paragraph{Metrics}

\paragraph{Competitive systems}
We implement four competitive ED systems, and three of them are based on variants of our proposed SGTB algorithm.\footnote{Our implementations are based on the scikit-learn package~\cite{pedregosa2011scikit}.} \emph{Gradient tree boosting} is a local model that employs only local features to make independent decisions for every entity mention. Note that our local model is different from that presented by~\newcite{yamada2016joint}, where they treat ED as binary classification for each mention-entity pair. \emph{SGTB-BS} is a Structured Gradient Tree Boosting model trained with Beam Search with early update strategy. \emph{SGTB-BSG} uses Beam Search with Gold path training strategy presented in~\autoref{sec:model:bsg}. Finally, \emph{SGTB-BiBSG} exploits Bidirectional Beam Search with Gold path to leverage information from both past and future for better local search. 

In addition, we compare against best published results on all the datasets. To ensure fair comparisons, we group results according to candidate selection system that different ED systems adopted.

\paragraph{Parameter tuning}

We tune all the hyperparameters on the AIDA-dev set. We use recommended hyperparameter values from scikit-learn to train regression trees, except for the maximum depth of the tree, which we choose from $\{3,5,8\}$. After a set of preliminary experiments, we select the beam size from $\{3,4,5,6\}$.  The best values for the two hyperparameters are $3$ and $4$ respectively. As mentioned in~\autoref{sec:model}, the learning rate is set to $1$. We train SGTB for at most $500$ epochs (i.e., fit at most $500$ regression trees). During training, we check the performance on the development set every $25$ epochs to perform early stopping.  Training takes $3$ hours for SGTB-BS and SGTB-BSG, and takes $9$ hours for SGTB-BiBSG on 16 threads.

\subsection{Results}
\label{sec:exp:results}

\begin{table} [t!]
\centering
\small
\addtolength{\tabcolsep}{-2pt}
\begin{tabular}{lcc}
    \toprule
    System  & PPRforNED & In-KB acc.  \\ \midrule
    \multicolumn{3}{l}{\it Published results} \\[2pt]
    \newcite{lazic2015plato} & &  86.4 \\
    \newcite{huang2015leveraging} & &  86.6 \\
    \newcite{chisholm2015entity} & &  88.7 \\
    \newcite{ganea2016probabilistic} & &  87.6 \\
    \newcite{guo2016robust} & & 89.0 \\
    \newcite{globerson2016collective} & & 91.0 \\
    \newcite{yamada2016joint} & & 91.5 \\
    \newcite{ganea2017deep} & & 92.2 \\[5pt]
    \multicolumn{3}{l}{\it Our implementations} \\[2pt]
    Gradient tree boosting & & 88.4 \\
    SGTB-BS & & 91.7 \\
    SGTB-BSG & & 92.4 \\
    SGTB-BiBSG & & \textbf{93.0} \\ \midrule
    \multicolumn{3}{l}{\it Published results} \\[2pt]
    \newcite{pershina2015personalized} & \checkmark & 91.8 \\
    \newcite{yamada2016joint} & \checkmark & 93.1 \\
    \newcite{sil2018neural} & \checkmark & 94.0 \\[5pt]
    \multicolumn{3}{l}{\it Our implementations} \\[2pt]
    Gradient tree boosting & \checkmark & 93.1 \\
    SGTB-BS & \checkmark & 95.1 \\
    SGTB-BSG & \checkmark & 95.5 \\
    SGTB-BiBSG & \checkmark & \textbf{95.9} \\
    \bottomrule
\end{tabular}
\caption{In-domain evaluation: in-KB accuracy results on the AIDA-test set. Checked PPRforNED indicates that the system uses PPRforNED~\cite{pershina2015personalized} to select candidate entities.The best results are in \textbf{bold}.}
\label{tab:aida}
\end{table}

\begin{table*} [t!]
\centering
\small
%\addtolength{\tabcolsep}{-2pt}
\begin{tabular}{lccccc}
    \toprule
    System & AQUAINT & MSNBC & ACE & CWEB & WIKI  \\ \midrule
    \multicolumn{6}{l}{\it Published results} \\[2pt]
    \newcite{fang2016entity} & 88.8 & 81.2 & 85.3 & - & - \\
    \newcite{ganea2016probabilistic} & 89.2 & 91.0 & 88.7 & - & - \\
    \newcite{milne2008learning} & 85.0 & 78.0 & 81.0 & 64.1 & 81.7 \\
    \newcite{hoffart2011robust} & 56.0 & 79.0 & 80.0 & 58.6 & 63.0 \\
    \newcite{ratinov2011local} & 83.0 & 75.0 & 82.0 & 56.2 & 67.2 \\
    \newcite{cheng2013relational} & 90.0 & 90.0 & 86.0 & 67.5 & 73.4 \\
    \newcite{guo2016robust} & 87.0 & 92.0 & 88.0 & 77.0 & \textbf{84.5} \\
    \newcite{ganea2017deep} & 88.5 & \textbf{93.7} & 88.5 & 77.9 & 77.5 \\ [5pt]
    \multicolumn{6}{l}{\it Our implementations} \\[2pt]
    Gradient tree boosting & 90.3 & 91.1 & \textbf{89.2} & 78.8 & 75.0 \\
    SGTB-BS & \textbf{90.5} & 92.4 & 88.9 & 81.7 & 76.4 \\
    SGTB-BSG & 89.4 & 92.5 & 88.6 & 81.7 & 78.4 \\
    SGTB-BiBSG & 89.9 & 92.6 & 88.5 & \textbf{81.8} & 79.2 \\
    \bottomrule
\end{tabular}
\caption{Cross-domain evaluation: Bag-of-Title (BoT) F1 results on ED datasets. The best results are in \textbf{bold}.}
\label{tab:wned}
\end{table*}

\paragraph{In-domain results} 
In-domain evaluation results are presented in~\autoref{tab:aida}. As shown, SGTB achieves much better performance than all previously published results. Specifically, SGTB-BiBSG outperforms the previous SOTA system~\cite{ganea2017deep} by $0.8\%$ accuracy, and improves upon the best published results when employing the PPRforNED candidate selection system by $1.9\%$ accuracy. Global information is clearly useful, as it helps to boost the performance by $2$-$4$ points of accuracy, depending on the candidate generation system. In terms of beam search training strategies, BiBSG consistently outperforms BSG and beam search with early update. By employing more point-wise functional gradients to train the regression trees and leveraging global information from both past and future to carry on local search, BiBSG is able to find better global solutions than alternative training strategies.  

\paragraph{Cross-domain results} 
As presented in~\autoref{tab:wned}, cross-domain experimental results are a little more mixed. SGTB-BS and SGTB-BSG perform quite competitively compared with SGTB-BiBSG. In a cross-domain evaluation setting, the test data is drawn from a different distribution as the training data. Therefore, less expressive models may be preferred as they may learn more abstract representations that will generalize better to out-of-domain data. Nevertheless, our SGTB models achieve better performance than best published results on three of the five popular ED datasets. Specifically, SGTB-BS outperforms the prior SOTA system by absolute $4\%$ F1 on the CWEB dataset, and SGTB-BiBSG performs consistently well across different datasets.  %Training and testing SGTB models in an in-domain evaluation setting for these datasets may require annotation of new data, which we leave for future work. 

% ***********************************************************
% Related Work
\section{Related work}
\label{sec:related}

\paragraph{Entity disambiguation}

Most ED systems consist of a local component that models relatedness between a mention and a candidate entity, as well as a global component that produces coherent entity assignments for all mentions within a document. Recent research has largely focused on joint resolution of entities, which is usually performed by maximizing the global
topical coherence between entities. As discussed above, directly optimizing the coherence objective is computationally intractable, and several heuristics and approximations have been proposed to address the problem. \newcite{hoffart2011robust} use an iterative heuristic to remove unpromising mention-entity edges. \newcite{yamada2016joint} employ a two-stage approach, in which global information is incorporated in the second stage based on local decisions from the first stage. Approximate inference techniques have been widely adopted for ED. \newcite{cheng2013relational} use an integer linear program (ILP) solver. Belief propagation (BP) and its variant loopy belief propagation (LBP) have been used by~\newcite{ganea2016probabilistic} and~\newcite{ganea2017deep} respectively. We employ another standard approximate inference algorithm, beam search, in this work. To make beam search a better fit for SGTB training, we propose BiBSG that improves beam search training on stability and effectiveness.

\paragraph{Structured gradient tree boosting}

Gradient tree boosting has been used in some of the most accurate systems for a variety of classification and regression problems~\cite{babenko2011robust,wu2010adapting,yamada2016joint}. However, gradient tree boosting is seldom studied in the context of structured learning, with only a few exceptions. \newcite{dietterich2004training} propose TreeCRF that replaces the linear scoring function of a CRF with a scoring function given by a gradient tree boosting model. TreeCRF achieves comparable or better results than CRF on some linear chain structured prediction problems. \newcite{bagnell2007boosting} extend the Maximum Margin Planning (MMP; Ratliff et al., 2006\nocite{ratliff2006maximum}) algorithm to structured prediction problems by learning new features using gradient boosting machines. \newcite{yang2015smart} present a general SGTB framework that is flexible in the choice of loss functions and specific structures. They also apply SGTB to the task of tweet entity linking with a special non-overlapping structure. By decomposing the structures into local substructures, exact inference is tractable in all the aforementioned works. Our work shows that we can train SGTB models efficiently and effectively even with approximate inference. This extends the utility of SGTB models to a wider range of interesting structured prediction problems.

% ***********************************************************
% Conclusion and Future Work
\section{Conclusion and future work}
\label{sec:con}

In this paper, we present a structured gradient tree boosting model for entity disambiguation. Entity coherence modeling is challenging, as exact inference is prohibitively expensive due to the pairwise entity relatedness terms in the objective function. We propose an approximate inference algorithm, BiBSG, that is designed specifically for SGTB to solve this problem. Experiments on benchmark ED datasets suggest that the expressive SGTB models are extremely good at dealing with the task of ED. SGTB significantly outperforms all previous systems on the AIDA-CoNLL dataset, and it also achieves SOTA results on many other ED datasets even in the cross-domain evaluation setting. SGTB is a family of structured learning algorithms that can be potentially applied to other core NLP tasks. In the future, we would like to investigate the effectiveness of SGTB on other information extraction tasks, such as relation extraction and coreference resolution.

% ***********************************************************
% Acknowledgments
\section{Acknowledgments}

We thank Prabhanjan Kambadur and other people in the Bloomberg AI team for their valuable comments on earlier version of this paper. We also thank the NAACL reviewers for their helpful feedback. This work also benefitted from discussions with Mark Dredze and Karl Stratos.

% ***********************************************************
\bibliographystyle{acl}
\bibliography{cite-strings,cites,cite-definitions}

\end{document}